%% file: Vollbeitrag.tex
\documentclass{MechatronikTagung}

\usepackage[english]{babel}	
\usepackage[utf8]{inputenc} % Deutsche Sonderzeichen/Umlaute gestatten
\usepackage[T1]{fontenc}
\usepackage{amsmath} % Zur Abbildung mathematischer Symbole und Formeln
\usepackage{amssymb} 
\usepackage{graphicx}
\usepackage{overpic}
\usepackage{booktabs} % Erweiterte Tabellenumgebung
\usepackage{cite} % Erweitertes Zitiermöglichkeiten
\usepackage{array,contour}% http://ctan.org/pkg/{array,contour}
\usepackage{mathtools}
\usepackage{xfrac}
\usepackage{paralist}
\usepackage{xcolor}
\usepackage[hidelinks]{hyperref}
\usepackage[ruled, linesnumbered]{algorithm2e}
\usepackage[caption=false, font=footnotesize]{subfig}
\usepackage{siunitx}

\newcommand{\mr}[1]{ \mathrm{#1} }
\newcommand{\bs}[1]{ \contour[3]{black}{$#1$} }
\makeatletter
\newcommand{\removelatexerror}{\let\@latex@error\@gobble}
\makeatother

\definecolor{klemmungfarbe}{RGB}{124,60,70}
\definecolor{kollisionfarbe}{RGB}{184,122,50}
\definecolor{unzul}{RGB}{230,60,36}
\definecolor{interakt}{RGB}{0,255,50}
\definecolor{install}{RGB}{174,74,132}

\graphicspath{{./Images/}}

\titeldeutsch{Optimierte parallelkinematische Roboter für die Mensch-Roboter-Kollaboration durch kombinierte Struktur- und Maßsynthese}

\titelenglisch{Towards Optimized Parallel Robots for Human-Robot Collaboration by Combined Structural and Dimensional Synthesis}

\autorenliste{%
Aran Mohammad, Thomas Seel, and Moritz Schappler\\%; \quad  \\[6pt]
Leibniz University Hannover, Institute of Mechatronic Systems, 30823 Garbsen, Germany\\%; \quad 
Contact: aran.mohammad@imes.uni-hannover.de
}

\kurzfassungdeutsch{%in maximal 10 Zeilen
Parallelkinematische Roboter bieten aufgrund ihrer geringeren bewegten Massen und den damit erlaubten höheren Geschwindigkeiten ein Potential für die Mensch-Roboter-Kollaboration (MRK).
Jedoch steigen die Kollisions- und Klemmgefahren aufgrund der parallelen Beinketten. 
Diese Gefahren werden im vorliegenden Beitrag kinematisch und kinetostatisch modelliert, um sie anschließend als Zielfunktionen durch eine kombinierte Struktur- und Maßsynthese in einer Partikelschwarmoptimierung zu minimieren.
Neben den Klemmgefahren innerhalb und zwischen benachbarten kinematischen Ketten wird die Rückttreibbarkeit quantifiziert, um eine Detektierbarkeit theoretisch zu garantieren. 
Als eine weitere MRK-relevante Zielfunktion wird der größte Eigenwert der im Arbeitsraum formulierten Massenmatrix minimiert, um Auswirkungen von Kollisionen zu reduzieren.
Die multikriterielle Optimierung führt zu verschiedenen Pareto-optimalen  Roboterstrukturen. 
Die Ergebnisse zeigen, dass durch die Optimierung eine deutliche Verbesserung der MRK-Merkmale möglich ist und eine Hexa-Struktur (6-\underline{R}US) hinsichtlich der Zielfunktionen und aufgrund ihrer einfacheren Gelenkstruktur zu favorisieren ist. 
}

\kurzfassungenglisch{%
Parallel robots (PR) offer potential for human-robot collaboration (HRC) due to their lower moving masses and higher speeds. 
However, the parallel leg chains increase the risks of collision and clamping. 
In this work, these hazards are described by kinematics and kinetostatics models to minimize them as objective functions by a combined structural and dimensional synthesis in a particle-swarm optimization. 
In addition to the risk of clamping within and between kinematic chains, the back-drivability is quantified to theoretically guarantee detectability via motor current. 
Another HRC-relevant objective function is the largest eigenvalue of the mass matrix formulated in the operational-space coordinates to consider collision effects. 
Multi-objective optimization leads to different Pareto-optimal PR structures. 
The results show that the optimization leads to significant improvement of the HRC criteria and that a Hexa structure (6-\underline{R}US) is to be favored concerning the objective functions and due to its simpler joint structure. 
}

\begin{document}

%Reviewerkommentare:
%\begin{itemize}
%    \item im Bereich der Systemoptimierung der Roboterkinematik --> Was sind die Vorteile unseres Konzeptes?
%    \item Schränkt die Sicherheitsoptimierung die Dynamik des Roboters ein? 
%    \item Ist die optimale PKM auch für andere Trajektorien optimal?
%    \item Nicht nur von der AMUN-PKM ausgehen, sondern einen größeren PKM-Raum beachten
%    \item 
%\end{itemize}

\section{Introduction}\label{sec:1_einleitung}
    Human-robot collaboration (HRC) enhances flexibility and productivity for non-fully automated tasks by combining the advantages of human and robotic-assisted work. 
    Human safety is ensured in a physical interaction by the safety mechanisms integrated into the collaborative robot (cobot) by monitoring the force in the event of contact. 
    Threshold values of contact forces and pressures are specified to avoid biomechanical stresses ~\cite{iso15066}, which are based on an energetic description of the contact depending on the affected human-body region. 
    
    The energy transferred to the human in the collision case results from the moving masses and the robot's speed projected onto the contact location. 
    Compared to industrial robots, serial cobots use lightweight designs as one possibility for reducing mass. 
    Another approach is using parallel robots (PR), typically characterized by fixed actuators connected via in-parallel kinematic chains to a mobile platform ~\cite{Merlet.2006}.
    The lighter moving masses of the passive chains allow higher speeds of the PR while maintaining the same energy thresholds. 
    However, the parallel structure of the kinematic chains increases the risk of collisions and, in particular, clamping. 
    Unintended contacts such as collisions and clamping must be considered regardless of the kinematic structure.

    \subsection{State of the Art}	
        Methods for contact detection can be based on \emph{exteroceptive}~\cite{Merckaert.2022} or \emph{proprioceptive}~\cite{Haddadin.2017} information.
        Image-based methods sense exteroceptive information, allowing its integration into robot path planning to avoid or at least handle contacts before their actual occurrence.
        However, sensors for joint-angle and motor-torque measurement are advantageous for rapid contact due to higher sample frequencies. 
        Based on the measurements, model-based disturbance observers can be used to estimate external forces that enable the \emph{detection}, \emph{localization} and \emph{identification} of accidental contacts.
    	Modeling inaccuracies, such as those for friction effects, lead to false-positive or false-negative detection, making it necessary to perform a component-wise threshold value comparison of the estimated external forces. 
    	Learning algorithms may also be applied for contact detection using estimated external joint torques by incorporating physical knowledge via a momentum observer~\cite{Lim.2021}.
    	
    	Depending on the result of the contact detection, reactions are triggered, for instance, by switching to an admittance controller~\cite{Luca.2006}. 
    	Once the contact localization is completed, an interaction control regarding the contact location can be implemented~\cite{Magrini.2014}, which enables manual reconfiguration.
    	
        \begin{figure}[t!]
    		\vspace{-1.5mm}			
    		\centering
    		\includegraphics[width=\columnwidth]{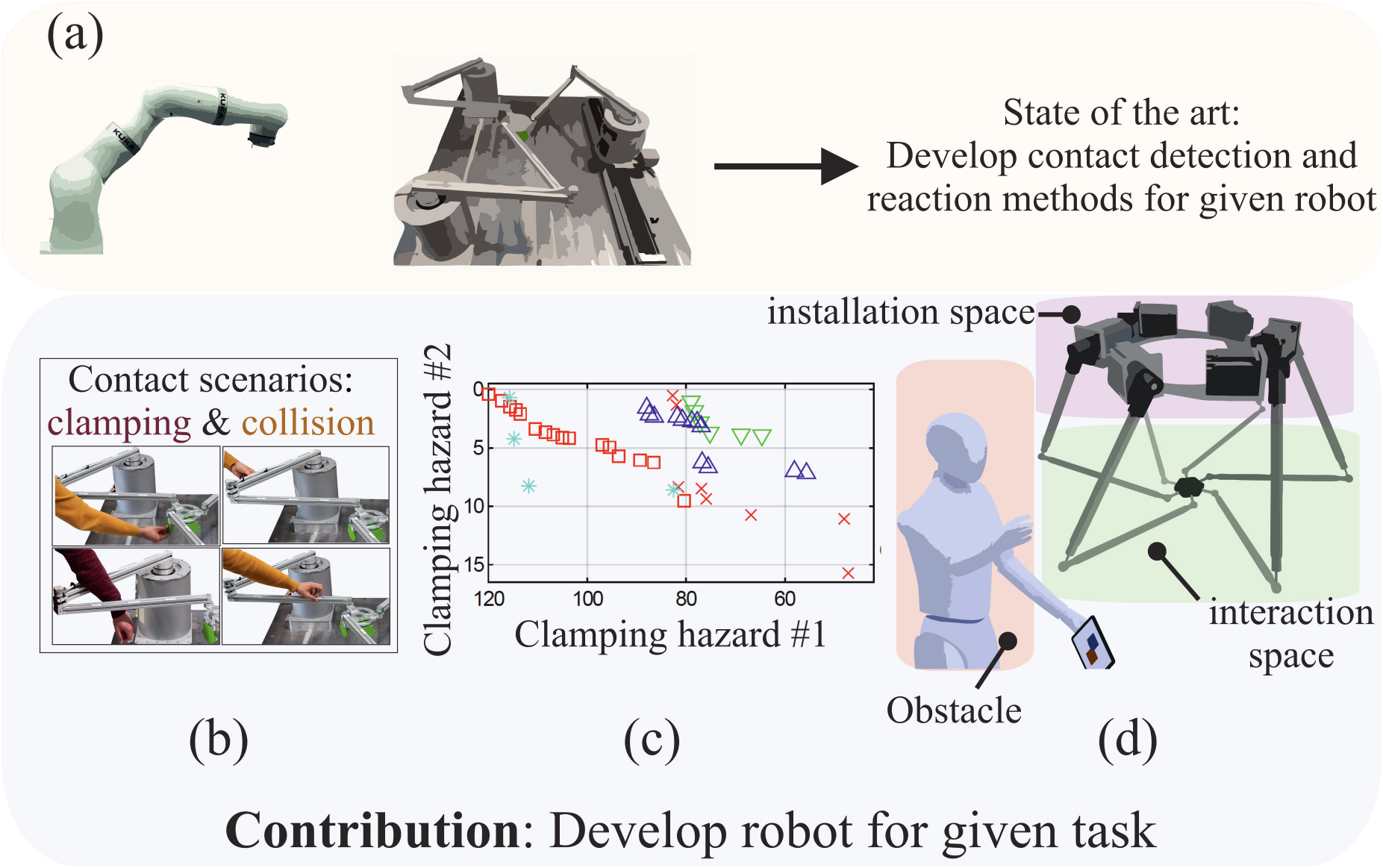}
    		\captionof{figure}{(a) Serial and parallel robot. The contribution of this work are: (b) \textcolor{klemmungfarbe}{Clamping}- and \textcolor{kollisionfarbe}{collision} scenarios, (c) clamping hazards in a Pareto diagram, and (d) optimal parallel robot (symbolic image) with minimal clamping hazards, which complies with the constraints of the impermissible space, the planned interaction space, and the allowed installation space.}
    		\label{fig:synt_mrkpkm}
    		\vspace{-1.5mm}
    	\end{figure}
    	
    	Methods for \emph{contact detection and reaction on PR} are demonstrated in the authors' previous works~\cite{Mohammad.2023, Mohammad.2023_IsolLoc, Mohammad.2023_Reaction, Mohammad.2023_UncQuant}. 
    	Contacts on a planar PR are detected with a momentum observer to remove collision and clamping contacts using reaction strategies such as a retraction movement or opening a clamping gap.

    	The previous studies demonstrate methods on \emph{existing} serial and parallel robots. 
    	However, the hazards arising from \emph{collision} and, in particular, \emph{clamping} contacts can already be \emph{incorporated into the design phase}.
    	
    	The design of the structure can be realized by specifying the joint type and arrangement (\emph{structural synthesis}) and the geometric parameters between the joints (\emph{dimensional synthesis})~\cite{Merlet.2006}, based on the optimization of one or more objective functions. 
        % In~\cite{Sterneck.2023} a PR optimized for automated mobile weed control is introduced by minimizing the objective function which incorporates the drive torques and speeds.
        % Constraints to maintain joint-angle limits and avoid self-collisions restrict the optimization problem, solved by a particle swarm optimization.
        Since, for parallel robots, the latter has a strong influence on the robot performance~\cite{Merlet.2006}, both synthesis steps should be combined, as introduced by~\cite{Krefft2006}, and formalized in~\cite{SchapplerOrt2020} into a larger optimization scheme with the attempt to incorporate all relevant constraints and objectives using multi-objective particle swarm optimization.
        For instance, in~\cite{Sterneck.2023}, the scheme was applied to optimize a PR  for automated mobile weed reduction by minimizing the drive torques and speeds with several constraints regarding installation space, accuracy and self-collision.

        %\begin{itemize}
         % \item MRK-bezogene Arbeiten, wie unsere Vorarbeiten, Maßsynthese-PKM
          %\item Ein Absatz zur Maßsynthese-MRK: Recherche nach weiteren Arbeiten 
        %\end{itemize}
	\subsection{Contribution of this Work}
	    The previous section shows that the \emph{consideration of clamping hazards in the design phase} of a PR represents a research gap that is addressed by this work with the contribution illustrated in Fig.~\ref{fig:synt_mrkpkm}:
	    \begin{compactitem}
	        \item Kinetostatic and kinematic objective functions are derived to describe clamping hazards in a human-robot collaboration with parallel robots.
	        \item A kinetostatic projection of contact forces onto the drive torques is performed to evaluate the detectability of external contact forces according to~\cite{iso15066} in the motor current.
	        \item The dimensional synthesis is performed by optimizing the objective functions for a reference trajectory and reference points. 
	        %These objective functions of the resulting Pareto diagram represent clamping hazards in one (via the clamping angle) or between two kinematic chains (via the collision distance). 
	        \item The discussion of results by Pareto diagrams and radar charts gives directions for the detailed construction of the HRC-optimized PR.
	    \end{compactitem}
        The paper has the following structure.
		Section~\ref{sec:2_methoden} summarizes the notation and kinematics modeling from the previous work~\cite{Mohammad.2023}.
		The contribution of this work is introduced in Sections~\ref{ssec:guetefunktional} and~\ref{ssec:optimierung}
		%\ref{sec:3_dimsyn} 
		by formulating and solving the constrained optimization problem. 
		A use case with specified requirements is presented in Section~\ref{sec:4_anwendungsszenario} with results following in Section~\ref{sec:5_ergebnisse}. Section~\ref{sec:6_zusammenfassung} concludes the paper.
\section{Kinematics and Kinetostatics of a Parallel Robot in Contact}\label{sec:2_methoden}
	A parallel robot with~$n$ platform degrees of freedom and~$n$ kinematic leg chains is considered (fully parallel). 
	The coordinates of the end effector and the coordinates of the active and passive joints are denoted by~$\bs{x}{\in}\mathbb{R}^n, \bs{q}_\mr{a}{\in}\mathbb{R}^n$, and $\bs{q}_\mr{p}$.
	All joint coordinates are summarized by the vector~$\bs{q}{\in}\mathbb{R}^{n_{q}}$ with~$n_{q}{=}n(n+\mathrm{dim}(\bs{q}_\mr{p}))$.
	By closing vector loops~\cite{Merlet.2006}, the full and reduced kinematic constraints
	\begin{subequations}\label{eq:KinCon}
		\begin{align}
			\bs{\delta}(\bs{x},\bs{q})&=\bs{0},\\
			\bs{\delta}_\mr{red}(\bs{x},\bs{q}_\mr{a})&=\bs{0}
		\end{align} 
	\end{subequations} 
	are formed. 
	Using the full and reduced constraints, the active and passive joint angles of a given pose are calculated, which corresponds to the inverse kinematics 
	\begin{equation}\label{eq:InvKin}
		\bs{q}{=}\bs{\mr{IK}}(\bs{x}).
	\end{equation}
	
	Differential kinematics are defined by the equations
	\begin{subequations}\label{eq:DifKin_Jac}
		\begin{align}
			\dot{\bs{q}}&={-}\bs{\delta}_{\partial \bs{q}}^{-1} \enskip \bs{\delta}_{\partial \bs{x}} \enskip \dot{\bs{x}}=\bs{J}_{q,x}\dot{\bs{x}},\\
			%%%%%%%%%%%%%%%%%%%%
			\dot{\bs{x}}&= {-}\left(\bs{\delta}_{\mr{red},\partial \bs{x}}\right)^{-1} (\bs{\delta}_{\mr{red},\partial \bs{q}_\mr{a}}) 	\dot{\bs{q}}_\mr{a}=\bs{J}_{x,q_\mr{a}}\dot{\bs{q}}_\mr{a}
		\end{align}
	\end{subequations} 
	with the notation~$\bs{a}_{\partial \bs{b}}{\coloneqq} \sfrac{\partial \bs{a}}{\partial \bs{b}}$ and the Jacobian matrices\footnote{For reasons of readability, the dependencies of~$\bs{q}$ and~$\bs{x}$ are omitted.}~$\bs{J}_{q, x}{\in}\mathbb{R}^{n_{q} \times n}$ and~$\bs{J}_{x, q_\mr{a}}{\in}\mathbb{R}^{n\times n}$.
	From a force and moment~$\bs{F}$ expressed in the operational space coordinates, the projected joint torque~$\bs{\tau}{=}\bs{J}_{x,q_\mr{a}}^\mr{T}\bs{F}$ is then determined by kinetostatics according to the principle of virtual work.
	
    \begin{figure*}[h]
        \graphicspath{{./Images/}}
        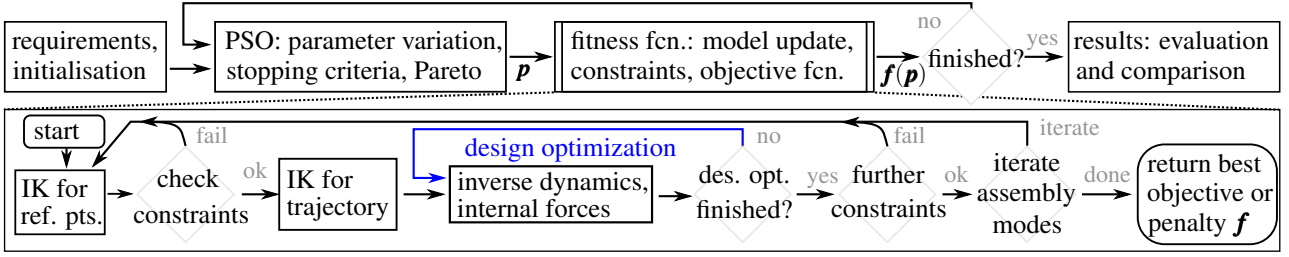
        \vspace{-2mm}
        \captionof{figure}{Overall procedure for the dimensional synthesis of a robot using  hierarchical constraints, mod. from~\cite{SchapplerOrt2020}}
        \vspace{-2mm}
        \label{fig:optimierungsschema}
    \end{figure*}
    
	The kinematics is now extended by the relationship of an \textit{arbitrary} point~$\mr{C}$ on the~$i$-th leg chain to the actuated joint coordinates by describing the desired pose~$\bs{x}_\mr{C}(\bs{q}_i, \bs{q}_j)$ using the~$j$-th kinematic chain in addition to the representation with the joint coordinates~$\bs{q}_i$ of leg-chain~$i$.
	This corresponds to closing vector loops via the~$i$-th and~$j$-th chain at the point~$\mr{C}$. 
	If the representation is set up for each leg chain,~$\bs{x}_\mr{C}(\bs{q})$ follows, which results in the differential relationship 
	\begin{equation}
	    \dot{\bs{x}}_\mr{C}{=}\bs{J}_{x_\mr{C},q} \dot{\bs{q}}.
	    \label{eq:DifKin}
	\end{equation}
	Including~(\ref{eq:DifKin_Jac}), it follows
	\begin{subequations}\label{eq:contactJacobian} 
		\begin{align}
			\dot{\bs{x}}_\mr{C} &= \bs{J}_{x_\mr{C},q} \dot{ \bs{q}}\\
			&=\bs{J}_{x_\mr{C},q} \bs{J}_{q,x} \dot{\bs{x}} = \bs{J}_{x_\mr{C}, x} \dot{\bs{x}} \\ 
			&= \bs{J}_{x_\mr{C}, x} \bs{J}_{x, q_\mr{a}} \dot{\bs{q}}_\mr{a} = \bs{J}_{x_\mr{C}, q_\mr{a}} \dot{\bs{q}}_\mr{a}
		\end{align} 
	\end{subequations}
	with the Jacobian matrices~$\bs{J}_{x_\mr{C}, x}$ and $\bs{J}_{x_\mr{C}, q_\mr{a}}$.
	With the principle of virtual work, the kinetostatic projection 
	\begin{equation} \label{eq:KinetostaticProj}
    	\bs{\tau}_\mr{ext}{=}\bs{J}_{x_\mr{C}, q_\mr{a}}^\mr{T} \bs{F}_\mr{ext}
    \end{equation}
	between an external contact wrench~$\bs{F}_\mr{ext}{=}(\bs{f}^\mr{T}, \bs{m}^\mr{T})^\mr{T}$ and the drive torques~$\bs{\tau}_\mr{ext}$ is obtained.
	
	Furthermore, the kinetostatic projection is suitable for evaluating the back-drivability of a PR.
	With an identified friction model for the drives and threshold values for maximum permissible contact forces~\cite{iso15066}, the kinetostatic projection yields the theoretically possible contact detection via motor current measurements.

%\section{Dimensional Synthesis}\label{sec:3_dimsyn} %  regarding Clamping Hazards		
%    This section begins with the presentation of the clamping hazards as objective functions (Sec.~\ref{ssec:guetefunktional}), followed by the formulation of the optimization problem method in Sec.~\ref{ssec:optimierung}.

    \section{Objective Functions for HRC} %the representation of clamping hazards}
    \label{ssec:guetefunktional}
		HRC-related criteria are represented by four objectives~$f_i$. % $f_1$ -- $f_4$.
		
		The criterion~$f_1$ according to Algorithm~\ref{alg:reactMotPl} accounts for \emph{detection of critical quasi-stationary contacts on the human hand}~\cite{iso15066}. % with $||\bs{f}_{\mr{ext}}|| {=} \SI{140}{\newton}$ and $||\bs{m}_{\mr{ext}}|| {=} \SI{0}{\newton\metre}$.
		Three points are sampled along every leg-chain segment with radial forces sampled by \SI{15}{\degree}. 
		Force directions on the platform are distributed evenly in a half-sphere by 200 samples.
		Low values of the resulting actuator torque are critical for the detection. 
		They are used as a worst case over all trajectory samples as maximization criterion, see line~\ref{alg:reactMotPl_f3calc} in Alg.~\ref{alg:reactMotPl}.		

		With~(\ref{eq:InvKin}) the \emph{angles of all passive joints} are known and the minimum angle over all joints and the reference trajectory is selected as the second objective function~$f_2= q_\mr{p,min}$ to account for the risk of clamping.
		
	    The lowest distances of any segment to the non-adjacent segments (on other kinematic chains) for all trajectory samples form the \emph{second clamping criterion}, resulting in the maximization objective~$f_3{=}d_{\min}$. %the minimum value $d_{\min}$ as the third objective function .
		Capsule segments (cylinders and half-spheres) are assumed to use simple geometry for distance calculation.
		A safety distance to the platform is kept to avoid finding the shortest distance only at the platform attachment.
		This assumes additional safety measures around the platform.
		
		The effective mass~\cite{iso15066} is calculated as the collision-relevant minimization objective~$f_4$ as the largest eigenvalue of the translational part of the robot's mass matrix in platform coordinates over the trajectory.
		The eigenvalues are considered to map contacts of all spatial directions implicitly.
		
        \begin{algorithm}[b] \nonumber
            \caption{Calculating the criterion~$f_1$}
           		\label{alg:reactMotPl}
                {\small 
                    \SetKwInOut{Input}{Input}
                    \SetKwInOut{Output}{Output}
                    \Input{$n_\mr{T}$ (number of trajectory samples), $n_\mr{l}$ (test locations on all bodies of a robot), $n_\mr{d}$ (directions)} % , $m$ (number of motors/legs)
                    \Output{Lowest possible external actuator torque}
                    $f_1\gets\infty$ \tcp*[f]{initialize the objective function} \\
                    \For(\tcp*[f]{iterate joint configuration}){$i_q{=}1$ to $n_\mr{T}$} 
                        {
                        \For(\tcp*[f]{iterate location}){$i_\mr{l}{=}1$ to $n_\mr{l}$} 
                            {
                            \For(\tcp*[f]{iterate direction}){$i_\mr{d}{=}1$ to $n_\mr{d}$} 
                                {
          							$\bs{\tau}_\mr{ext}{:=}\bs{J}_{x_\mr{C}, q_\mr{a}}^\mr{T} \bs{F}_\mr{ext}$ \tcp*[f]{(\ref{eq:KinetostaticProj}) with $||\bs{f}_{\mr{ext}}|| {=} \SI{140}{\newton}$} % $}
          							$f_1 \gets \mathrm{min}(f_1, \mathrm{max}(|\bs{\tau}_\mr{ext}|))$ \label{alg:reactMotPl_f3calc}			
                                }
                            }
                        }
                }
        \end{algorithm}
        
		The HRC criteria are only evaluated in a \emph{defined interaction space}, see Fig.~\ref{fig:synt_mrkpkm}(d).
		The required maximum drive torques for the trajectory movement~$f_5$ and the maximum drive speeds~$f_6$ are used as further minimization criteria to reduce the dimensioning of the drives and thereby the robot's costs.

    \section{Optimization Problem}\label{ssec:optimierung}
    To obtain the HRC-optimized parallel robot, a combined structural and dimensional synthesis of all possible robots' geometric parameters $\bs{p}$ w.r.t. multiple objectives $\bs{f}$ from Sec.~\ref{ssec:guetefunktional} is performed.
    The 11 to 17 parameters are, e.g., the overall scaling parameter, leg length and joint alignment (using Denavit-Hartenberg notation), base and platform size, or the distance between joint pairs.
    The overall approach is sketched in Fig.~\ref{fig:optimierungsschema} and uses multi-objective particle-swarm optimization (PSO) and follows a concept termed ``hierarchical constraints'' \cite{SchapplerOrt2020}, which includes constraints (self-collision, installation space, joint-angle limits) as penalties in the fitness function (increasing with severity). 
    The evaluation aborts upon violation of a constraint to save computation time.
    	\begin{figure*}[t]
        	\vspace{-2mm}
        	\centering
        	\begin{overpic}%[,grid,tics=10]
        	{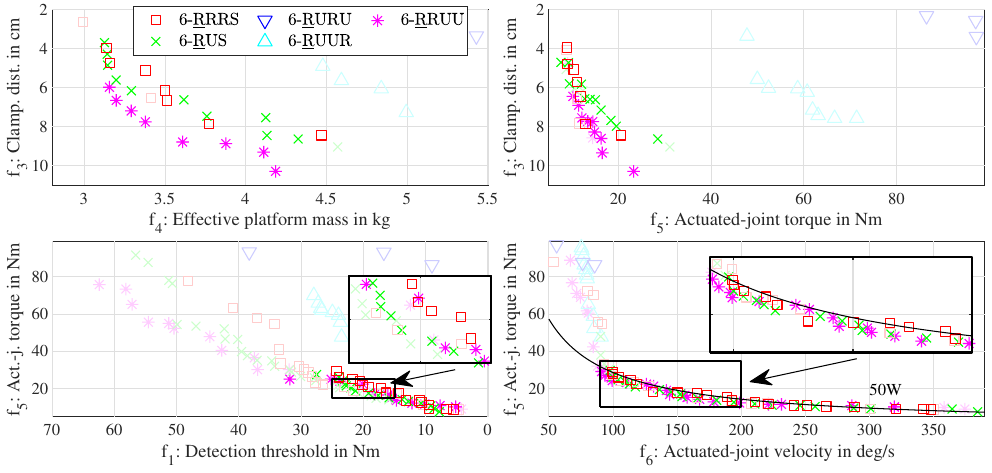}
        	\put(0,2){\textbf{(c)}}
        	\put(50,2){\textbf{(d)}}
        	\put(0,25){\textbf{(a)}}
        	\put(50,25){\textbf{(b)}}
        	\end{overpic}
        	\captionof{figure}{Pareto diagrams of the results with only Pareto-dominant solutions regarding the respective criteria. Solutions exceeding the thresholds on optimization criteria are plotted transparently. Good values are located in the lower left of the diagrams.}
        	\label{fig:ergebnisse}
        \end{figure*}
        \begin{figure*}[t]
    		\vspace{-1.5mm}	
        	\centering
        	\graphicspath{{./Images/}}
        	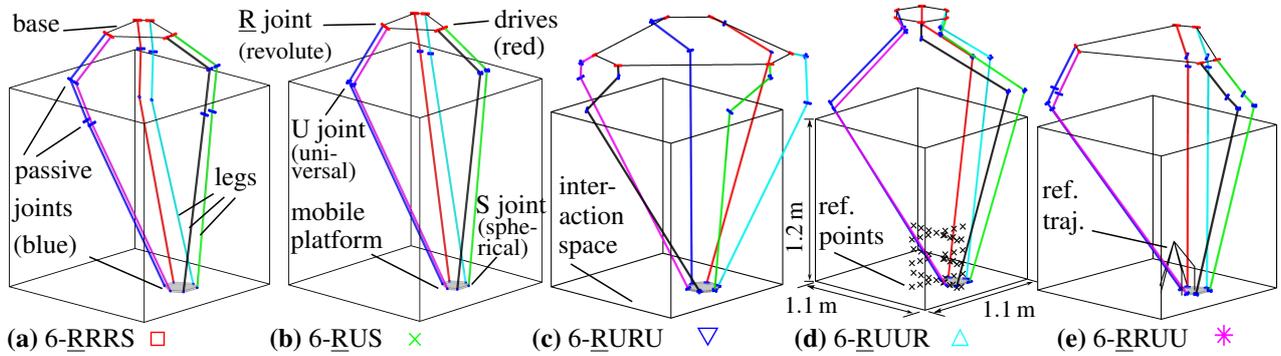    	\captionof{figure}{Kinematic sketches of parallel robots with different leg chains, based on Fig.~\ref{fig:ergebnisse}(a); performance shown in Fig.~\ref{fig:radarchart}(a) }
        	\label{fig:ergebnisse_roboter}
    		\vspace{-1.5mm}	
        \end{figure*}
                \begin{figure*}[htb]
    		\vspace{-1.5mm}			
    		\centering
    		\includegraphics[width=\textwidth]{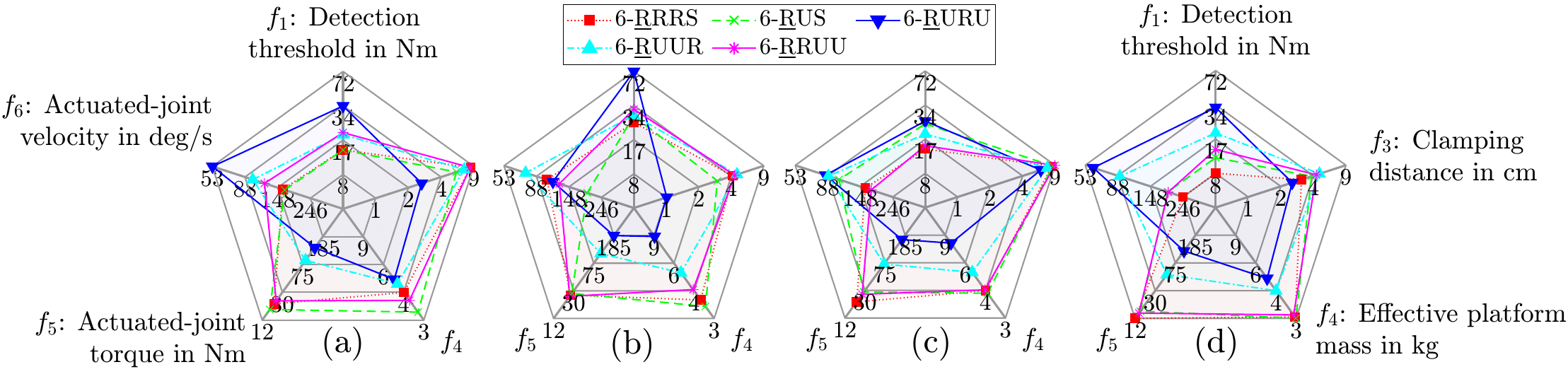}
    		\captionof{figure}{Radar charts of the five objective functions with the five parallel-robot structures: (a) Solutions from Fig.~\ref{fig:ergebnisse_roboter} as compromise of~$f_3$ and~$f_4$, (b) from the respective maximum of~$f_1$, (c) maximum of~$f_3$, and (d) minimum of ~$f_4$; best values are on the outside.}
    		\label{fig:radarchart}
    		\vspace{-1.5mm}
    	\end{figure*}
    The optimization of the different robots (by leg-chain type and permutations of base- and platform-coupling-joint alignments) is done in parallel on the LUH's computing cluster using a \textsc{Matlab} implementation, available as open source \cite{GitHub_StructDimSynth}.
    The inverse kinematics (\ref{eq:InvKin}) is first applied to several reference points and then using (\ref{eq:DifKin}) on a reference trajectory.
    If internal forces are exceeded in a lightweight construction assuming aluminum tubes for the robot structure, an additional design optimization based on a single-objective PSO increases the strength and diameter of the structure with possible negative effects on mass, actuator load and self-collision, used as objective/constraints. 

    The computation is performed for different assembly modes, e.g., elbow-inside assembly mode to reduce the installation space, or alternating elbow inside/outside assembly to increase the clamping distance. % between leg chains.
    
\section{Use Case: Pick and Place}\label{sec:4_anwendungsszenario}
    A pick-and-place task with a full-mobility PR ($n{=}6$) and revolute actuation at the base is considered as a benchmark, oriented at the motion used in the case study of \cite[p.\,155ff.]{Krefft2006} for the assembly of light bulbs.
    The task includes reference points with constant orientation in a cylinder with $r${=}\SI{250}{\milli\metre} and $h${=}\SI{350}{\milli\metre}, and a reference trajectory between two pick poses in the back of the workspace and four assembly poses around the center with tilting of \SI{15}{\degree} and rotation of \SI{20}{\degree}.
    An additional platform mass of \SI{2}{\kilo\gram} is assumed to account for a gripper and/or handling object.
    
    As sketched in Fig.~\ref{fig:synt_mrkpkm}(d), a cuboid interaction space of \SI{1.1}{\metre}$\times$\SI{1.1}{\metre}$\times$\SI{1.2}{\metre} is defined within the allowed installation space of the PR. 
    The latter is defined by an area of \SI{1.0}{\metre}$\times$\SI{1.0}{\metre} starting at a table height of \SI{0.8}{\metre} and a bigger cuboid for the base with \SI{2.0}{\metre}$\times$\SI{2.0}{\metre}$\times$\SI{1.0}{\metre} above the human when standing at the table (above \SI{2.1}{\metre}).
    Constraints on performance criteria are summarized in Table~\ref{tab:opt_lim}.
    If a criterion is not subject to optimization (rows \ref{mrkconstr:poserr}, \ref{mrkconstr:condj}, and \ref{mrkconstr:matstress}), a violation leads to a penalty and discarding the particle.
    Otherwise (rows \ref{mrkconstr:exttorque}, \ref{mrkconstr:clampangle}, \ref{mrkconstr:clampdist}, and~\ref{mrkconstr:actforce}), it is kept in the optimization to increase the population diversity and excluded in the evaluation of Sec.~\ref{sec:5_ergebnisse}.
    %The objective functions are evaluated in the interaction space. 
    %If a particle in the optimization leads to a solution which violates the limits specified in Table~\ref{tab:opt_lim}, it is declared invalid and the particle is discarded. 

    \newcounter{mrkconstraints}
    \begin{table}[htb]
        	\caption{Limits of HRC and PR-related target values}
        	\vspace{-1.5mm}
        	\label{tab:opt_lim}
        	\begin{center}
        		\begin{tabular}{|c|c|c|}
        			\hline
        			\multicolumn{3}{|c|}{\textsc{HRC-related objective functions}} \\
        			\hline \refstepcounter{mrkconstraints}\themrkconstraints\label{mrkconstr:exttorque} & 	lowest allowed external torque &$>\SI{5}{\newton\metre}$\\
        			\hline
        		    \refstepcounter{mrkconstraints}\themrkconstraints\label{mrkconstr:clampangle} &	minimum clamping angle&$>\SI{30}{\degree}$\\
        			\hline
        			\refstepcounter{mrkconstraints}\themrkconstraints\label{mrkconstr:clampdist} &   minimum clamping distance&$>\SI{3}{\centi\metre}$\\
        			\hline
        			\hline
        			\multicolumn{3}{|c|}{\textsc{Limits of PR-related properties}}\\
        			\hline
        			\refstepcounter{mrkconstraints}\themrkconstraints\label{mrkconstr:actforce}  &  max. actuated-joint torque&$<\SI{30}{\newton\metre}$\\
        			\hline
        			\refstepcounter{mrkconstraints}\themrkconstraints\label{mrkconstr:poserr} &  largest theoretical position error&$<\SI{500}{\micro\metre}$\\
        			\hline
        			\refstepcounter{mrkconstraints}\themrkconstraints\label{mrkconstr:condj}  &  condition number of $\bs{J}_{x,q_\mr{a}}$&$<500$\\
        			\hline
        			 \refstepcounter{mrkconstraints}\themrkconstraints\label{mrkconstr:matstress} &  material stress& $<$ 50\%\\
        			\hline
        		\end{tabular}
        	\end{center}
        	\vspace{-1.5mm}	
        \end{table}

    %\begin{itemize}
        %\item Anforderungen einer auszudenkenden Pick\&Place: Erreichbarkeit von Eckpunkten, woraus eine Trajektorie hervorgeht. Aufgabe ist 3T3R. Aus praktischen Gründen werden PKM mit 3 oder vier %Gelenken pro Beinkette geprüft
        %\item Wie wird die Simulation umgesetzt?
        %\item Fiktives MRK-Szenario mit PKM. Bild mit Menschen/PKM und verschiedenen Räumen (Bauraum, Arbeitsraum, Kollaborationsraum). Im Kollaborationsraum werden die Grenzen betrachtet, im %Arbeitsraum nicht, weswegen im Letzteren räumlich getrennt sind 
        %\item Grenzen bezüglich MRK-Zielfunktionen: Klemmabstand~$>\SI{30}{\milli\metre}$, Klemmwinkel~$>\SI{30}{\degree}$, Kontaktkraft-Antriebsmoment~$>\SI{5}{\newton\metre}$
        %\item Grenzen bezüglich physikalischer Leistungsmerkmale: Antriebsmoment~$<\SI{30}{\newton\metre}$, Präzision (theoretischer Positionsfehler)~$<\SI{500}{\micro\metre}$, %Jacobimatrix-Konditionszahl~$<500$, Materialspannung<50\,\%
%    \end{itemize}
%    \begin{figure}[h]
        %Schematisches Bild der Aufgabe als 3D-Plot. Verschiedene Quader zeigen Bauraum, Interaktionsraum und Trajektorie sowie Mensch und PKM. Benutze Rendering basierend auf Arbeitsraum aus cds\_show\_task.
        %ToDo Aran: Geht das mit vertretbarem Aufwand? Evtl. als Teaser-Bild direkt vorne auf dem Paper. Vorlage: https://seafile.projekt.uni-hannover.de/smart-link/d4035f80-9837-47f5-9a45-bd1252156696/
        %    	\captionof{figure}{Platzhalter: Bild der Aufgabe}
        %    	\label{fig:aufgabe}
    %\end{figure}

\section{Results}\label{sec:5_ergebnisse}

    The optimization results are shown in Fig.~\ref{fig:ergebnisse} by four two-dimensional Pareto diagrams regarding the optimization criteria $f_i$ from Sec.~\ref{ssec:guetefunktional}.
    All five principally possible leg chains provide feasible results regarding technical constraints.
    The structures 6-\underline{R}URU and {6-\underline{R}UUR} are infeasible for realization due to their high required actuator torques.
    One possible parameterization for each robot is shown as a kinematic sketch in Fig.~\ref{fig:ergebnisse_roboter}, chosen according to the normalized and equally-weighted performance for the clamping and collision criteria in Fig.~\ref{fig:ergebnisse}(a).
    
    The Pareto diagrams provide observations that can be used to select and dimension a robot structure.
    A strong correlation between the effect of external forces~$f_1$ to the required maximal actuator torque~$f_5$ is visible in Fig.~\ref{fig:ergebnisse}(c).
    The platform weight dominates the dynamics effects and has the same influence on the actuators as the external force.
    To allow \emph{sensitive collision detection} already for low external forces, \emph{high actuated-joint torques will result}, which requires higher gear rations to reduce the motor torques to omit large and expensive motors.
    However, the gear dimensioning should reduce gear friction by choosing the minimum number of gear stages, which increases the detection threshold.
    All Pareto-dominant solutions in Fig.~\ref{fig:ergebnisse}(d) require a rated power of the actuators of about \SI{50}{\watt}, which shows that increasing the actuator-joint torque by longer lever arms equally lowers the actuated-joint velocity.
    Solutions with higher actuator torques than about \SI{30}{\newton\metre} require a stronger structural dimension due to material stress limits (checked in the design optimization), which increases the self-weight and thereby further increases the required torque.
    
    The trade-off between clamping distance and effective mass relevant for collision severity becomes clear from Fig.~\ref{fig:ergebnisse}(a).
    A possible explanation is that long limbs or a large platform increase the clamping distance but also the mass.
    The latter effect increases the required actuator torque, explaining the trade-off in Fig.~\ref{fig:ergebnisse}(b).
    Regarding these criteria, the 6-\underline{R}RUU slightly outperforms the 6-\underline{R}RRS and 6-\underline{R}US.
    
    The \emph{clamping-angle criterion} $f_2$ was removed from the evaluation as all structures received the optimal value of \SI{180}{\degree} by moving the passive joints above the interaction space, which does not deteriorate other criteria.
    However, in runs of the optimizations with smaller allowed installation space, this was not possible and a general task-independent statement can not be derived.

	\begin{figure}[t]
		\centering
		\graphicspath{{./Images/}}
		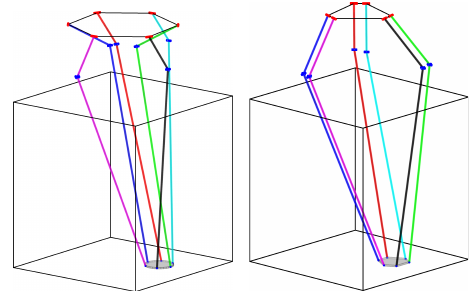
		\captionof{figure}{Different variants of the Hexa structure: (a) mixed elbow inside/outside configuration and (b) optimized for clamping distance}
		\label{fig:skizze_ellenbogenhexaabwechselnd}
	\end{figure}
        \begin{figure}[t]
    		\vspace{-6mm}	% großer Abstand, wenn Fig.6 und 7 direkt übereinander sind		
    		\centering
    		\includegraphics[width=\columnwidth]{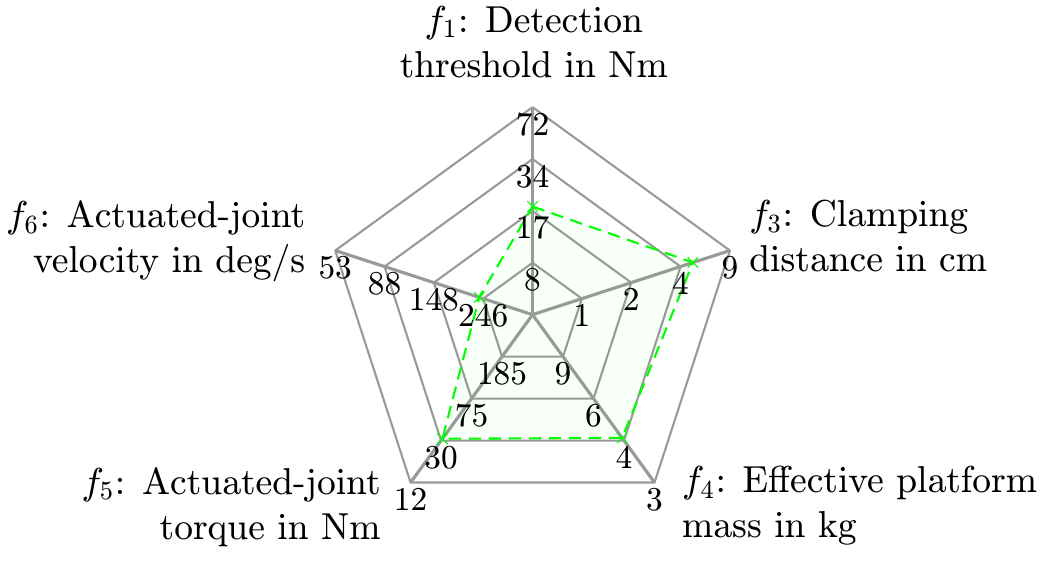}
    		\vspace{-6mm}
    		\captionof{figure}{Radar chart for the 6-\underline{R}US structure from Fig.~\ref{fig:skizze_ellenbogenhexaabwechselnd}(a) with mixed elbow inside/outside configuration}
    		\label{fig:radarchart_single}
    		\vspace{-1.5mm}
    	\end{figure}

    Since the overall Pareto diagrams do not provide a dominating structure, a further inspection of the single possible parameterizations has to be done, using the radar charts in Fig.~\ref{fig:radarchart} to show the mutual influence of the multiple objectives which can not be shown in the 2D Pareto diagrams.
    %Further, from the diagrams it is not clear if Pareto-optimal sets in one of the diagrams also is Pareto-dominant in the others.
    % Due to the five criteria, the Pareto diagrams do not provide a clearly dominant
    %Therefore, single parameterizations are shown in a radar chart in Fig.~\ref{fig:radarchart}
    %
    Again, 6-\underline{R}RUU, 6-\underline{R}RRS, and 6-\underline{R}US provide acceptable performance.
    Since clamping hazards were identified as one of the significant challenges in HRC with parallel robots, the achieved distances of at least about \SI{9}{\centi\metre} when \emph{optimizing for clamping} in Fig.~\ref{fig:radarchart}(c) provide the strongest argument for a safety-based design decision.
    This addresses the most relevant scenario of clamping the users' hands.
    When optimized for clamping via $f_3$, the Hexa robot, depicted in Fig.~\ref{fig:skizze_ellenbogenhexaabwechselnd}(b), outperforms the other structures regarding the detection threshold ($f_1$) and has equal effective mass ($f_4$).
    Also, from the perspective of design, assembly, and operation, this presents the favored solution due to the lower number of joints and links and fewer associated difficulties regarding tolerances, calibration, and changing the working mode.

    The selected optimal results did not contain mixed assembly modes (elbow inside and outside), depicted in Fig.~\ref{fig:skizze_ellenbogenhexaabwechselnd}(a), which was assumed to increase the collision distance and simplify reaching through the robot.
    A possible model-inherent reason lies in the computation of the distance measure, which is dominated by the smallest distance of the leg-chain collision bodies near the platform.
    The overall performance of this structure is however still good, as visible in Fig.~\ref{fig:radarchart_single}.
    These mixed-mode solutions were more frequent in the Pareto front for a smaller installation space, similar to \cite{Sterneck.2023}, where a small installation space lead to an unconventional elbow-inside assembly.

\section{Summary}\label{sec:6_zusammenfassung}

The developed safety criteria for parallel robots in human-robot collaboration are applied as objectives and constraints in synthesizing optimized kinematic structures.
Due to the multitude of criteria to be considered, many variants of three different general structures reach the requirements.
The selection is done based on the clamping-distance criterion, which was the most important.
The established Hexa structure is favored due to its simple design.
%Before approaching a detailed construction, questions regarding different assembly modes and a re-adjustment of the distance criterion will be considered.
As a next step, the detailed construction is approached by dimensioning the drives and selecting or constructing appropriate joints from catalogs.
Further runs of the dimensional synthesis may be necessary for the latter.

\section*{Acknowledgement}
    The authors acknowledge the support by the German Research Foundation (DFG) under grant numbers 444769341 and~341489206.

% TODO: Lieber bib-Datei?
\bibliographystyle{spmpsci_unsrt}
\bibliography{literatur} % literatur.bib

\clearpage  % Verhindert automatischen Umbruch auf der letzten Seite. Bei Bedarf entfernen.
\end{document}

%% file: Images/optimization_flowchart.pdf_tex
%% Creator: Inkscape 1.2.1 (9c6d41e410, 2022-07-14), www.inkscape.org
%% Accompanies image file 'optimization_flowchart.pdf' (pdf, eps, ps)
\begingroup%
  \makeatletter%
  \providecommand\color[2][]{%
    \errmessage{(Inkscape) Color is used for the text in Inkscape, but the package 'color.sty' is not loaded}%
    \renewcommand\color[2][]{}%
  }%
  \providecommand\transparent[1]{%
    \errmessage{(Inkscape) Transparency is used (non-zero) for the text in Inkscape, but the package 'transparent.sty' is not loaded}%
    \renewcommand\transparent[1]{}%
  }%
  \providecommand\rotatebox[2]{#2}%
  \newcommand*\fsize{\dimexpr\f@size pt\relax}%
  \newcommand*\lineheight[1]{\fontsize{\fsize}{#1\fsize}\selectfont}%
  \ifx\svgwidth\undefined%
    \setlength{\unitlength}{476.22047244bp}%
    \ifx\svgscale\undefined%
      \relax%
    \else%
      \setlength{\unitlength}{\unitlength * \real{\svgscale}}%
    \fi%
  \else%
    \setlength{\unitlength}{\svgwidth}%
  \fi%
  \global\let\svgwidth\undefined%
  \global\let\svgscale\undefined%
  \makeatother%
  \begin{picture}(1,0.20238095)%
    \lineheight{1}%
    \setlength\tabcolsep{0pt}%
    \put(0.00637473,0.16163113){\color[rgb]{0,0,0}\makebox(0,0)[lt]{\lineheight{1.25}\smash{\begin{tabular}[t]{l}requirements,\end{tabular}}}}%
    \put(0.17334628,0.16163113){\color[rgb]{0,0,0}\makebox(0,0)[lt]{\lineheight{1.25}\smash{\begin{tabular}[t]{l}PSO: parameter variation,\end{tabular}}}}%
    \put(0.17428735,0.13806842){\color[rgb]{0,0,0}\makebox(0,0)[lt]{\lineheight{1.25}\smash{\begin{tabular}[t]{l}stopping criteria, Pareto\end{tabular}}}}%
    \put(0.83968412,0.13806842){\color[rgb]{0,0,0}\makebox(0,0)[lt]{\lineheight{1.25}\smash{\begin{tabular}[t]{l}and comparison\end{tabular}}}}%
    \put(0,0){\includegraphics[width=\unitlength,page=1]{optimization_flowchart.pdf}}%
    \put(0.11991614,0.05286575){\color[rgb]{0,0,0}\makebox(0,0)[lt]{\lineheight{1.25}\smash{\begin{tabular}[t]{l}check\end{tabular}}}}%
    \put(0.1014003,0.02410493){\color[rgb]{0,0,0}\makebox(0,0)[lt]{\lineheight{1.25}\smash{\begin{tabular}[t]{l}constraints\end{tabular}}}}%
    \put(0,0){\includegraphics[width=\unitlength,page=2]{optimization_flowchart.pdf}}%
    \put(0.71548649,0.17776157){\color[rgb]{0.58823529,0.58823529,0.58823529}\makebox(0,0)[lt]{\lineheight{0}\smash{\begin{tabular}[t]{l}\small{no}\end{tabular}}}}%
    \put(0.00697469,0.13806842){\color[rgb]{0,0,0}\makebox(0,0)[lt]{\lineheight{1.25}\smash{\begin{tabular}[t]{l}initialisation\end{tabular}}}}%
    \put(0,0){\includegraphics[width=\unitlength,page=3]{optimization_flowchart.pdf}}%
    \put(0.44411676,0.16163113){\color[rgb]{0,0,0}\makebox(0,0)[lt]{\lineheight{1.25}\smash{\begin{tabular}[t]{l}fitness fcn.: model update, \end{tabular}}}}%
    \put(0.44411676,0.13806842){\color[rgb]{0,0,0}\makebox(0,0)[lt]{\lineheight{1.25}\smash{\begin{tabular}[t]{l}constraints, objective fcn.\end{tabular}}}}%
    \put(0.40124248,0.14066793){\color[rgb]{0,0,0}\makebox(0,0)[lt]{\lineheight{0}\smash{\begin{tabular}[t]{l}\small{$\bs{p}$}\end{tabular}}}}%
    \put(0,0){\includegraphics[width=\unitlength,page=4]{optimization_flowchart.pdf}}%
    \put(0.01384525,0.04427905){\color[rgb]{0,0,0}\makebox(0,0)[lt]{\lineheight{1.25}\smash{\begin{tabular}[t]{l}IK for\end{tabular}}}}%
    \put(0.01328278,0.02301788){\color[rgb]{0,0,0}\makebox(0,0)[lt]{\lineheight{1.25}\smash{\begin{tabular}[t]{l}ref. pts.\end{tabular}}}}%
    \put(0,0){\includegraphics[width=\unitlength,page=5]{optimization_flowchart.pdf}}%
    \put(0.66427411,0.05601555){\color[rgb]{0,0,0}\makebox(0,0)[lt]{\lineheight{1.25}\smash{\begin{tabular}[t]{l}further\end{tabular}}}}%
    \put(0.647833,0.02871706){\color[rgb]{0,0,0}\makebox(0,0)[lt]{\lineheight{1.25}\smash{\begin{tabular}[t]{l}constraints\end{tabular}}}}%
    \put(0,0){\includegraphics[width=\unitlength,page=6]{optimization_flowchart.pdf}}%
    \put(0.36110985,0.07722248){\color[rgb]{0,0,1}\makebox(0,0)[lt]{\lineheight{1.25}\smash{\begin{tabular}[t]{l}design optimization\end{tabular}}}}%
    \put(0,0){\includegraphics[width=\unitlength,page=7]{optimization_flowchart.pdf}}%
    \put(0.89577346,0.06431587){\color[rgb]{0,0,0}\makebox(0,0)[lt]{\lineheight{1.25}\smash{\begin{tabular}[t]{l}return best\end{tabular}}}}%
    \put(0.89616821,0.01785618){\color[rgb]{0,0,0}\makebox(0,0)[lt]{\lineheight{0}\smash{\begin{tabular}[t]{l}penalty $\bs{f}$\end{tabular}}}}%
    \put(0,0){\includegraphics[width=\unitlength,page=8]{optimization_flowchart.pdf}}%
    \put(0.89545971,0.04142393){\color[rgb]{0,0,0}\makebox(0,0)[lt]{\lineheight{1.25}\smash{\begin{tabular}[t]{l}objective or\end{tabular}}}}%
    \put(0,0){\includegraphics[width=\unitlength,page=9]{optimization_flowchart.pdf}}%
    \put(0.02331902,0.09023038){\color[rgb]{0,0,0}\makebox(0,0)[lt]{\lineheight{1.25}\smash{\begin{tabular}[t]{l}start\end{tabular}}}}%
    \put(0,0){\includegraphics[width=\unitlength,page=10]{optimization_flowchart.pdf}}%
    \put(0.77515664,0.06476538){\color[rgb]{0,0,0}\makebox(0,0)[lt]{\lineheight{1.25}\smash{\begin{tabular}[t]{l}iterate\end{tabular}}}}%
    \put(0.76200683,0.04128202){\color[rgb]{0,0,0}\makebox(0,0)[lt]{\lineheight{1.25}\smash{\begin{tabular}[t]{l}assembly\end{tabular}}}}%
    \put(0.18751844,0.05940466){\color[rgb]{0.58823529,0.58823529,0.58823529}\makebox(0,0)[lt]{\lineheight{0}\smash{\begin{tabular}[t]{l}\small{ok}\end{tabular}}}}%
    \put(0.7367089,0.05651324){\color[rgb]{0.58823529,0.58823529,0.58823529}\makebox(0,0)[lt]{\lineheight{0}\smash{\begin{tabular}[t]{l}\small{ok}\end{tabular}}}}%
    \put(0,0){\includegraphics[width=\unitlength,page=11]{optimization_flowchart.pdf}}%
    \put(0.62881622,0.05544414){\color[rgb]{0.58823529,0.58823529,0.58823529}\makebox(0,0)[lt]{\lineheight{0}\smash{\begin{tabular}[t]{l}\small{yes}\end{tabular}}}}%
    \put(0.15055548,0.08710344){\color[rgb]{0.58823529,0.58823529,0.58823529}\makebox(0,0)[lt]{\lineheight{0}\smash{\begin{tabular}[t]{l}\small{fail}\end{tabular}}}}%
    \put(0.69693368,0.08710344){\color[rgb]{0.58823529,0.58823529,0.58823529}\makebox(0,0)[lt]{\lineheight{0}\smash{\begin{tabular}[t]{l}\small{fail}\end{tabular}}}}%
    \put(0.58982399,0.08710344){\color[rgb]{0.58823529,0.58823529,0.58823529}\makebox(0,0)[lt]{\lineheight{0}\smash{\begin{tabular}[t]{l}\small{no}\end{tabular}}}}%
    \put(0.77426461,0.0161199){\color[rgb]{0,0,0}\makebox(0,0)[lt]{\lineheight{1.25}\smash{\begin{tabular}[t]{l}modes\end{tabular}}}}%
    \put(0,0){\includegraphics[width=\unitlength,page=12]{optimization_flowchart.pdf}}%
    \put(0.81138183,0.09340305){\color[rgb]{0.58823529,0.58823529,0.58823529}\makebox(0,0)[lt]{\lineheight{0}\smash{\begin{tabular}[t]{l}\small{iterate}\end{tabular}}}}%
    \put(0,0){\includegraphics[width=\unitlength,page=13]{optimization_flowchart.pdf}}%
    \put(0.2217322,0.0517383){\color[rgb]{0,0,0}\makebox(0,0)[lt]{\lineheight{1.25}\smash{\begin{tabular}[t]{l}IK for\end{tabular}}}}%
    \put(0.22158522,0.03004957){\color[rgb]{0,0,0}\makebox(0,0)[lt]{\lineheight{1.25}\smash{\begin{tabular}[t]{l}trajectory\end{tabular}}}}%
    \put(0,0){\includegraphics[width=\unitlength,page=14]{optimization_flowchart.pdf}}%
    \put(0.35525024,0.05043408){\color[rgb]{0,0,0}\makebox(0,0)[lt]{\lineheight{1.25}\smash{\begin{tabular}[t]{l}inverse dynamics,\end{tabular}}}}%
    \put(0.35510326,0.02874536){\color[rgb]{0,0,0}\makebox(0,0)[lt]{\lineheight{1.25}\smash{\begin{tabular}[t]{l}internal forces\end{tabular}}}}%
    \put(0,0){\includegraphics[width=\unitlength,page=15]{optimization_flowchart.pdf}}%
    \put(0.84608636,0.05651324){\color[rgb]{0.58823529,0.58823529,0.58823529}\makebox(0,0)[lt]{\lineheight{0}\smash{\begin{tabular}[t]{l}\small{done}\end{tabular}}}}%
    \put(0.80077452,0.16467645){\color[rgb]{0.58823529,0.58823529,0.58823529}\makebox(0,0)[lt]{\lineheight{0}\smash{\begin{tabular}[t]{l}\small{yes}\end{tabular}}}}%
    \put(0,0){\includegraphics[width=\unitlength,page=16]{optimization_flowchart.pdf}}%
    \put(0.83799915,0.16163113){\color[rgb]{0,0,0}\makebox(0,0)[lt]{\lineheight{1.25}\smash{\begin{tabular}[t]{l}results: evaluation\end{tabular}}}}%
    \put(0.7235736,0.14738289){\color[rgb]{0,0,0}\makebox(0,0)[lt]{\lineheight{1.25}\smash{\begin{tabular}[t]{l}finished?\end{tabular}}}}%
    \put(0.68687692,0.13605744){\color[rgb]{0,0,0}\makebox(0,0)[lt]{\lineheight{0}\smash{\begin{tabular}[t]{l}\small{$\bs{f}(\bs{p})$}\end{tabular}}}}%
    \put(0,0){\includegraphics[width=\unitlength,page=17]{optimization_flowchart.pdf}}%
    \put(0.54417533,0.05554533){\color[rgb]{0,0,0}\makebox(0,0)[lt]{\lineheight{1.25}\smash{\begin{tabular}[t]{l}des. opt.\end{tabular}}}}%
    \put(0.54216057,0.02824684){\color[rgb]{0,0,0}\makebox(0,0)[lt]{\lineheight{1.25}\smash{\begin{tabular}[t]{l}finished?\end{tabular}}}}%
    \put(0,0){\includegraphics[width=\unitlength,page=18]{optimization_flowchart.pdf}}%
  \end{picture}%
\endgroup%

%% file: Images/Kinematik_Skizzen.pdf_tex
%% Creator: Inkscape 1.2.1 (9c6d41e410, 2022-07-14), www.inkscape.org
%% Accompanies image file 'Kinematik_Skizzen.pdf' (pdf, eps, ps)
\begingroup%
  \makeatletter%
  \providecommand\color[2][]{%
    \errmessage{(Inkscape) Color is used for the text in Inkscape, but the package 'color.sty' is not loaded}%
    \renewcommand\color[2][]{}%
  }%
  \providecommand\transparent[1]{%
    \errmessage{(Inkscape) Transparency is used (non-zero) for the text in Inkscape, but the package 'transparent.sty' is not loaded}%
    \renewcommand\transparent[1]{}%
  }%
  \providecommand\rotatebox[2]{#2}%
  \newcommand*\fsize{\dimexpr\f@size pt\relax}%
  \newcommand*\lineheight[1]{\fontsize{\fsize}{#1\fsize}\selectfont}%
  \ifx\svgwidth\undefined%
    \setlength{\unitlength}{476.22047244bp}%
    \ifx\svgscale\undefined%
      \relax%
    \else%
      \setlength{\unitlength}{\unitlength * \real{\svgscale}}%
    \fi%
  \else%
    \setlength{\unitlength}{\svgwidth}%
  \fi%
  \global\let\svgwidth\undefined%
  \global\let\svgscale\undefined%
  \makeatother%
  \begin{picture}(1,0.27380952)%
    \lineheight{1}%
    \setlength\tabcolsep{0pt}%
    \put(0,0){\includegraphics[width=\unitlength,page=1]{Kinematik_Skizzen.pdf}}%
    \put(0.62638881,0.0745913){\color[rgb]{0,0,0}\rotatebox{90}{\makebox(0,0)[lt]{\lineheight{1.25}\smash{\begin{tabular}[t]{l}\small{\SI{1.2}{\metre}}\end{tabular}}}}}%
    \put(0,0){\includegraphics[width=\unitlength,page=2]{Kinematik_Skizzen.pdf}}%
    \put(0.37220785,0.09066093){\color[rgb]{0,0,0}\makebox(0,0)[lt]{\lineheight{1.25}\smash{\begin{tabular}[t]{l}\small{(sphe-}\end{tabular}}}}%
    \put(0.37220785,0.07371147){\color[rgb]{0,0,0}\makebox(0,0)[lt]{\lineheight{1.25}\smash{\begin{tabular}[t]{l}\small{rical)}\end{tabular}}}}%
    \put(0,0){\includegraphics[width=\unitlength,page=3]{Kinematik_Skizzen.pdf}}%
    \put(0.00351997,0.00125762){\color[rgb]{0,0,0}\makebox(0,0)[lt]{\lineheight{1.25}\smash{\begin{tabular}[t]{l}\textbf{(a)} 6-\underline{R}RRS\end{tabular}}}}%
    \put(0.41464348,0.00125762){\color[rgb]{0,0,0}\makebox(0,0)[lt]{\lineheight{1.25}\smash{\begin{tabular}[t]{l}\textbf{(c)} 6-\underline{R}URU\end{tabular}}}}%
    \put(0.82576696,0.00125762){\color[rgb]{0,0,0}\makebox(0,0)[lt]{\lineheight{1.25}\smash{\begin{tabular}[t]{l}\textbf{(e)} 6-\underline{R}RUU\end{tabular}}}}%
    \put(0.62020522,0.00125762){\color[rgb]{0,0,0}\makebox(0,0)[lt]{\lineheight{1.25}\smash{\begin{tabular}[t]{l}\textbf{(d)} 6-\underline{R}UUR\end{tabular}}}}%
    \put(0.20908174,0.00125762){\color[rgb]{0,0,0}\makebox(0,0)[lt]{\lineheight{1.25}\smash{\begin{tabular}[t]{l}\textbf{(b)} 6-\underline{R}US\end{tabular}}}}%
    \put(0,0){\includegraphics[width=\unitlength,page=4]{Kinematik_Skizzen.pdf}}%
    \put(0.18489658,0.22902264){\color[rgb]{0,0,0}\makebox(0,0)[lt]{\lineheight{1.25}\smash{\begin{tabular}[t]{l}\small{(revolute)}\end{tabular}}}}%
    \put(0.00782937,0.25025315){\color[rgb]{0,0,0}\makebox(0,0)[lt]{\lineheight{1.25}\smash{\begin{tabular}[t]{l}base\end{tabular}}}}%
    \put(0.43596852,0.12180675){\color[rgb]{0,0,0}\makebox(0,0)[lt]{\lineheight{1.25}\smash{\begin{tabular}[t]{l}inter-\end{tabular}}}}%
    \put(0.43580857,0.09848755){\color[rgb]{0,0,0}\makebox(0,0)[lt]{\lineheight{1.25}\smash{\begin{tabular}[t]{l}action\end{tabular}}}}%
    \put(0.43547636,0.07518661){\color[rgb]{0,0,0}\makebox(0,0)[lt]{\lineheight{1.25}\smash{\begin{tabular}[t]{l}space\end{tabular}}}}%
    \put(0.18489658,0.25368871){\color[rgb]{0,0,0}\makebox(0,0)[lt]{\lineheight{1.25}\smash{\begin{tabular}[t]{l}\underline{R} joint\end{tabular}}}}%
    \put(0.2259249,0.16664262){\color[rgb]{0,0,0}\makebox(0,0)[lt]{\lineheight{1.25}\smash{\begin{tabular}[t]{l}U joint\end{tabular}}}}%
    \put(0.37063602,0.10671149){\color[rgb]{0,0,0}\makebox(0,0)[lt]{\lineheight{1.25}\smash{\begin{tabular}[t]{l}S joint\end{tabular}}}}%
    \put(0.38521039,0.25353368){\color[rgb]{0,0,0}\makebox(0,0)[lt]{\smash{\begin{tabular}[t]{l}drives\end{tabular}}}}%
    \put(0.22595917,0.1011008){\color[rgb]{0,0,0}\makebox(0,0)[lt]{\smash{\begin{tabular}[t]{l}mobile\end{tabular}}}}%
    \put(0.22575,0.07925151){\color[rgb]{0,0,0}\makebox(0,0)[lt]{\smash{\begin{tabular}[t]{l}platform\end{tabular}}}}%
    \put(0.38437372,0.23244693){\color[rgb]{0,0,0}\makebox(0,0)[lt]{\smash{\begin{tabular}[t]{l}(red)\end{tabular}}}}%
    \put(0,0){\includegraphics[width=\unitlength,page=5]{Kinematik_Skizzen.pdf}}%
    \put(0.76707724,0.02527445){\color[rgb]{0,0,0}\makebox(0,0)[lt]{\lineheight{1.25}\smash{\begin{tabular}[t]{l}\small{\SI{1.1}{\metre}}\end{tabular}}}}%
    \put(0,0){\includegraphics[width=\unitlength,page=6]{Kinematik_Skizzen.pdf}}%
    \put(0.0096938,0.13437786){\color[rgb]{0,0,0}\makebox(0,0)[lt]{\smash{\begin{tabular}[t]{l}passive\end{tabular}}}}%
    \put(0.00872179,0.10447915){\color[rgb]{0,0,0}\makebox(0,0)[lt]{\smash{\begin{tabular}[t]{l}joints\end{tabular}}}}%
    \put(0.00857414,0.07541341){\color[rgb]{0,0,0}\makebox(0,0)[lt]{\smash{\begin{tabular}[t]{l}(blue)\end{tabular}}}}%
    \put(0,0){\includegraphics[width=\unitlength,page=7]{Kinematik_Skizzen.pdf}}%
    \put(0.16619918,0.12866259){\color[rgb]{0,0,0}\makebox(0,0)[lt]{\lineheight{1.25}\smash{\begin{tabular}[t]{l}legs\end{tabular}}}}%
    \put(0.22605717,0.14889187){\color[rgb]{0,0,0}\makebox(0,0)[lt]{\lineheight{1.25}\smash{\begin{tabular}[t]{l}\small{(uni-}\end{tabular}}}}%
    \put(0.22605717,0.13430343){\color[rgb]{0,0,0}\makebox(0,0)[lt]{\lineheight{1.25}\smash{\begin{tabular}[t]{l}\small{versal)}\end{tabular}}}}%
    \put(0,0){\includegraphics[width=\unitlength,page=8]{Kinematik_Skizzen.pdf}}%
    \put(0.61211166,0.0279405){\color[rgb]{0,0,0}\makebox(0,0)[lt]{\lineheight{1.25}\smash{\begin{tabular}[t]{l}\small{\SI{1.1}{\metre}}\end{tabular}}}}%
    \put(0,0){\includegraphics[width=\unitlength,page=9]{Kinematik_Skizzen.pdf}}%
    \put(0.63924676,0.10634617){\color[rgb]{0,0,0}\makebox(0,0)[lt]{\lineheight{1.25}\smash{\begin{tabular}[t]{l}ref.\end{tabular}}}}%
    \put(0.63949284,0.08239721){\color[rgb]{0,0,0}\makebox(0,0)[lt]{\lineheight{1.25}\smash{\begin{tabular}[t]{l}points\end{tabular}}}}%
    \put(0,0){\includegraphics[width=\unitlength,page=10]{Kinematik_Skizzen.pdf}}%
    \put(0.81456287,0.11856139){\color[rgb]{0,0,0}\makebox(0,0)[lt]{\lineheight{1.25}\smash{\begin{tabular}[t]{l}ref.\end{tabular}}}}%
    \put(0.81480894,0.09461244){\color[rgb]{0,0,0}\makebox(0,0)[lt]{\lineheight{1.25}\smash{\begin{tabular}[t]{l}traj.\end{tabular}}}}%
  \end{picture}%
\endgroup%

%% file: Images/Kinematik_Skizzen2.pdf_tex
%% Creator: Inkscape 1.2.1 (9c6d41e410, 2022-07-14), www.inkscape.org
%% Accompanies image file 'Kinematik_Skizzen2.pdf' (pdf, eps, ps)
\begingroup%
  \makeatletter%
  \providecommand\color[2][]{%
    \errmessage{(Inkscape) Color is used for the text in Inkscape, but the package 'color.sty' is not loaded}%
    \renewcommand\color[2][]{}%
  }%
  \providecommand\transparent[1]{%
    \errmessage{(Inkscape) Transparency is used (non-zero) for the text in Inkscape, but the package 'transparent.sty' is not loaded}%
    \renewcommand\transparent[1]{}%
  }%
  \providecommand\rotatebox[2]{#2}%
  \newcommand*\fsize{\dimexpr\f@size pt\relax}%
  \newcommand*\lineheight[1]{\fontsize{\fsize}{#1\fsize}\selectfont}%
  \ifx\svgwidth\undefined%
    \setlength{\unitlength}{226.77165354bp}%
    \ifx\svgscale\undefined%
      \relax%
    \else%
      \setlength{\unitlength}{\unitlength * \real{\svgscale}}%
    \fi%
  \else%
    \setlength{\unitlength}{\svgwidth}%
  \fi%
  \global\let\svgwidth\undefined%
  \global\let\svgscale\undefined%
  \makeatother%
  \begin{picture}(1,0.6375)%
    \lineheight{1}%
    \setlength\tabcolsep{0pt}%
    \put(0,0){\includegraphics[width=\unitlength,page=1]{Kinematik_Skizzen2.pdf}}%
    \put(0.02363397,0.01127687){\color[rgb]{0,0,0}\makebox(0,0)[lt]{\lineheight{1.25}\smash{\begin{tabular}[t]{l}\textbf{(a)}\end{tabular}}}}%
    \put(0.53468872,0.01127687){\color[rgb]{0,0,0}\makebox(0,0)[lt]{\lineheight{1.25}\smash{\begin{tabular}[t]{l}\textbf{(b)}\end{tabular}}}}%
  \end{picture}%
\endgroup%